\documentclass[runningheads]{llncs}
\usepackage[T1]{fontenc}
\usepackage{graphicx,verbatim}
\usepackage{graphicx}
\usepackage{tabularx}
\usepackage{multirow}

\usepackage{amsmath}
\usepackage{color}

\usepackage{amsfonts} 
\usepackage[table]{xcolor}  
\usepackage[colorlinks=true, allcolors=blue]{hyperref}

\begin{document}
\title{Single Image Estimation of Cell Migration Direction by Deep Circular Regression}
\titlerunning{Single Image Estimation of Cell Migration Direction}

\author{Lennart Bruns\inst{1} \and
Lucas Lamparter \inst{2} \and
Milos Galic \inst{2} \and
Xiaoyi Jiang \inst{1}}

\authorrunning{L. Bruns et al.}
%
\institute{
$^1$ Faculty of Mathematics and Computer Science, University of Münster, Münster, Germany \\
$^2$ Institute of Medical Physics and Biophysics, University of Münster, Münster, Germany
}

\maketitle              
\begin{abstract}
In this paper, we address the problem of estimating the migration direction of cells based on a single image. A solution to this problem lays the foundation for a variety of applications that were previously not possible. To our knowledge, there is only one related work that employs a classification CNN with four classes (quadrants). However, this approach does not allow for detailed directional resolution. We tackle the single image estimation problem using deep circular regression, with a particular focus on cycle-sensitive methods. On two common datasets, we achieve a mean estimation error of $\sim\!17^\circ$, representing a significant improvement over previous work, which reported estimation error of $30^\circ$ and $34^\circ$, respectively.


\end{abstract}

\section{Introduction}

\begin{figure}[t]
    \centering
    \includegraphics[width=0.5\textwidth]{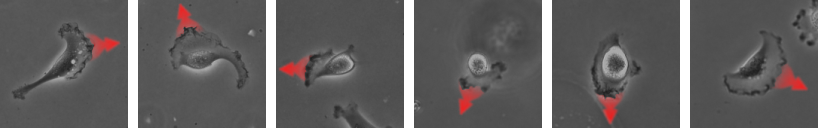} \\
    \includegraphics[width=0.5\textwidth]{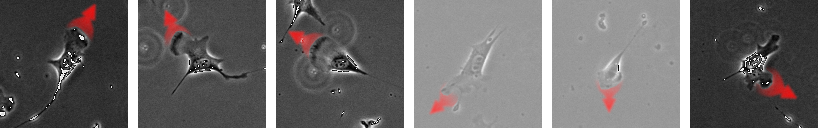} \\
    \hspace*{1mm}\includegraphics[width=0.5\textwidth]{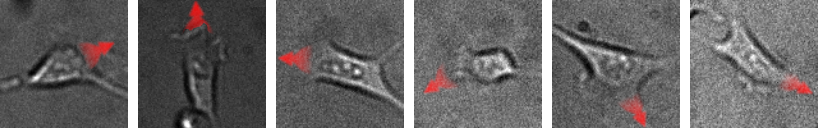}
    \vspace{-2mm}
    \caption{Example cell images with ground truth cell migration direction (red arrow). Top: NIH3T3 dataset. Middle: U373 dataset. Bottom: MS3T3 dataset.}
    \label{fig:examples}
\end{figure}

Microscopic images of individual cells are essential in biomedical research, with extracting meaningful information critical for addressing complex questions. Key tasks in cell image analysis include segmentation, classification, and tracking \cite{Liu2021,Xu2022}. This work focuses on single image estimation of cell migration direction (SIECMD), predicting the future direction of cell migration from a single image, as shown in Figure \ref{fig:examples}.

We employ deep learning, specifically deep circular regression, to learn shape features for SIECMD. To date, this problem has received limited attention. To our knowledge, the only related work is Nishimoto et al. \cite{Nishimoto2019}. Their method uses a 14-layer CNN (comprising eight convolutional layers, four max-pooling layers, and two fully connected layers) and formulates the task as a 4-class classification problem (upper left, upper right, lower left, lower right; one class per quadrant). We view this approach as a significant drawback since it offers only a coarse sampling of the full spectrum of directions on the unit circle, thereby restricting directional resolution. In contrast, our method addresses SIECMD through regression, with special attention to the circular nature of directional data.

Zhang’s thesis work \cite{Zhang2022} also addresses SIECMD, extending the approach of Nishimoto et al. \cite{Nishimoto2019} by using a variable number of classes -- rather than a fixed four -- that is tuned during training. However, this variable is determined using information from two consecutive images (at time points $t$ and $t+1$), which makes it unsuitable for a single image context. Furthermore, the estimated migration direction is not evaluated independently but is instead employed to support cell tracking in low-frame-rate videos. Another single image challenge is the classification of cell migration modes (continuous vs. discontinuous) \cite{GuptaL2020}.

Direction estimation has been explored in various contexts. For instance, wind direction estimation from SAR images \cite{Corazza2020} and the analysis of the angle of intrusion in porcine ventricular myocytes using Diffusion Tensor MRI \cite{Schmid2007} both rely on global analysis techniques, such as the Fourier transform, which fundamentally differ from the SIECMD problem. Similarly, while angle estimation is crucial for oriented object detection \cite{Wen2023}, that topic addresses a different scenario where the objects’ directions are clearly perceptible, e.g. in aerial imagery.

The remainder of the paper is organized as follows. In Section \ref{sec:siecmd}, we discuss the significance of the SIECMD problem. Section \ref{sec:methods} presents the details of our method, and Section \ref{sec:results} provides the experimental results and discussions. The paper concludes with further discussion in Section \ref{sec:conclusion}.

\section{SIECMD problem}
\label{sec:siecmd}

In vertebrates, 2D single-cell migration occurs in processes like Zebrafish germ cell migration \cite{Thebing2020}, epidermal keratocyte migration \cite{Labuz2022}, and leukocyte patrolling before transmigration \cite{Auffray2007,Honig2024}. While some systems are compatible with videomicroscopy \cite{Pewsey2021}, others, such as human samples, are less accessible for live-cell analysis. Estimating cell states (migration vs. stationary) from a single image could also aid in high-throughput drug screening. Additionally, migration data could be used in applications beyond biomedicine, such as cell simulation models like the Potts model (CPM) \cite{Graner1992,Wortel2021}. Replacing single-cell tracking with kinetic data from SIECMD could significantly boost throughput and quantitative accuracy in these analyses.

At the cellular level, single cell migration patterns are best described as intermittent migration \cite{Begemann2009,Maiuri2015}. Here, upon polarization, which may occur spontaneously or in response to a chemical cue, the cell extends protrusions in the future direction of migration \cite{Begemann2009,Jiang2005}. This polarized state, which remains stable for several minutes, yield a super-diffusive migration at length-scales relevant for cellular foraging \cite{Begemann2009,Maiuri2015}. Of importance for our work, these studies establish a clear relationship between the migration and shape of individual cells \cite{Moginer2009}.

\section{Method}
\label{sec:methods}

%
%

Circular data is measured on a circle in degrees or radians, which fundamentally differs from linear data. Due to its periodic nature (0° = 360°), traditional techniques developed for Euclidean space may not be directly applicable. The unique challenges of working with circular data have been widely acknowledged in the field of statistics \cite{Lee2010,Pewsey2021}, but not yet in computer vision (see \cite{Pan2024} for one of the few works in this area). Particularly, there is still a lack of standardized methods for designing deep CNN components specifically for circular regression.

To estimate the cell migration direction, several key details need to be specified, including direction encoding, activation functions, and loss functions. We explore various choices for each of these components. We introduce a small CNN to probe the best configuration of these choices. We then use this optimal configuration to fine-tune pre-trained large neural networks for the specific task at hand. Additionally, we study test-time augmentation as a means of further improving performance. These steps of our approach are detailed in the following.

\vspace{1mm}
\noindent{\bf Direction encoding.}
We consider two schemes for direction encoding:  1) an angle $\alpha \in [0, 2\pi)$.
2) a point on the unit circle $(x,y), \ x^2+y^2=1$. The choice of encoding also determines the number of output neurons (1 or 2) in the deep neural network for migration direction estimation.

\vspace{1mm}
\noindent{\bf Activation functions.}
ReLU is used as the activation function in all layers except for the last (output) layer. There, we define for the angle direction encoding the following activation function:
\begin{equation}
\label{eq:activation_mod}
\varphi_{cyclic}(x) \ = \ x \bmod 2\pi
\end{equation}
For the circle direction encoding, we consider two options: the identity function and the Sigmoid function:
\begin{equation}
\varphi_{identity}(x) \ = \ x; \quad
\varphi_{sigmoid}(x) \ = \ \frac{e^x-1}{e^x+1} \label{eq:activation-2N} 
\end{equation}
The modulo operator and Sigmoid function serve as activation functions for normalization, mapping values to $[0, 2\pi)$ and $[-1, 1]$, respectively.

\vspace{1mm}
\noindent{\bf Loss functions.}
For angle direction encoding, we define two loss functions, along with their quadratic variants, to compare the network's output and the ground truth:
\begin{eqnarray}
    \delta_{linear}(\alpha, \beta) &=& |\alpha-\beta|; \ 
    \delta_{linear}^2(\alpha, \beta)  = (\delta_{linear}(\alpha, \beta))^2 \label{eq_loss_linear} \\
    \delta_{cyclic}(\alpha, \beta) &=& \min(|\alpha - \beta|, 2\pi - |\beta - \alpha|); \ 
    \delta_{cyclic}^2(\alpha, \beta) = (\delta_{cyclic}(\alpha, \beta))^2 \label{eq_loss_cyclic} \\
    \delta_{cos}(\alpha, \beta) & = & -\cos(\alpha-\beta) \label{eq_loss_cos}
\end{eqnarray}
The loss functions (\ref{eq_loss_cyclic}) and (\ref{eq_loss_cos}) account for the circular nature of direction data, making them expected to perform better than the linear loss functions in (\ref{eq_loss_linear}). The loss function (\ref{eq_loss_cos}) has a mathematical interpretation as the maximum likelihood estimation for circular data, based on the von Mises distribution (see \cite{Bruns2024} for the derivation).

For the circle direction encoding we define three simple loss functions:
\begin{eqnarray}
\delta_{dist}((x_1,y_1), (x_2,y_2)) & = & |x_1-x_2|+|y_1-y_2| \\
\delta_{dist}^2((x_1,y_1), (x_2,y_2)) & = & (\delta_{dist}((x_1,y_1), (x_2,y_2)))^2 \\
\delta_{eucl}((x_1,y_1), (x_2,y_2)) &=& \sqrt{(x_1-y_2)^2+(x_2-y_2)^2}
\end{eqnarray}

\noindent{\bf Deep neural networks.}
We described several encoding schemes, activation functions (for the output layer), and loss functions, resulting in 21 possible combinations. We introduce a small CNN to probe the best configuration and then only use this optimal configuration to test the performance of larger networks. This probing CNN consists of two convolutional layers followed by max pooling, with a head made of three fully connected layers ending in one or two neurons, depending on the direction encoding (see Table \ref{tab:cnn}).

Fine-tuning pre-trained neural networks for a specific task reduces training effort and yields strong results \cite{Yu2022,Zhuang2021}. We explore several prominent neural network backbones (YOLOv8, ResNet50 \cite{He2016}, and EfficientNet \cite{Tan2021}). YOLOv8 was pre-trained on the COCO dataset, while ResNet50 and EfficientNet were pre-trained on ImageNet, providing valuable initial parameters for feature extraction. We selected the smallest architecture from each network and used only the backbone (up to the first fully connected layers) for our task, adapting the head to match the complexity of these architectures (see Table \ref{tab:cnn_backbone}). Since these backbones were pre-trained on well-established datasets, they required minimal retraining. The training process involved fine-tuning the entire model for 10 epochs, freezing the backbone weights, and then training the remaining layers for an additional 50 epochs. In Section \ref{sec:results}, we demonstrate that the EfficientNet backbone outperforms our simple CNN backbone, yielding the best results.

\begin{table}[t]
\centering
\begin{tabular}{l c r}
        \hline
        Layer & Output & \# Parameters \\
        \hline 
        \hline
        Input layer & (128, 128) & 0 \\         
        Conv2D (5x5), activation=ReLU & (124, 124, 16) & 416 \\  
        MaxPooling2D & (62, 62, 16) & 0 \\                                           
        Conv2D (3x3), activation=ReLU & (60, 60, 32) & 4640 \\      
        MaxPooling2D & (30, 30, 32) & 0 \\                                          
        Flatten & (28800) & 0 \\        
        Dense, activation=ReLU & (256) & 7373056 \\ 
        Dense, activation=ReLU & (16) & 4112 \\
        Dense, activation=$\varphi$ & ($\#$) & $\#$ \\        
        \hline
\end{tabular}
\caption{Probing CNN for finding the best configuration. Cell image size: $128 \times 128$. Symbol $\#$ indicates that the value there depends on the direction encoding.}
\label{tab:cnn}
\end{table}

\begin{table}[tb]
\centering
\begin{tabular}{l c r}
        \hline
        Layer & Output & \# Parameters \\
        \hline 
        \hline
        Backbone & & \\ 
        \hline 
        GlobalAveragePooling2D & (\#) & 0 \\        
        Dense, activation=ReLU & (1024) & \# \\ 
        Dense, activation=ReLU & (256) & 262400 \\
        Dense, activation=$\varphi$ & ($\#$) & $\#$ \\         
        \hline
\end{tabular}
\caption{Fine-tuning architecture. Backbone refers to the foreign model that is connected to our customized head with the fully connected layers. Symbol $\#$ indicates that the value there depends on the direction encoding or the size of the previous layers.}
\label{tab:cnn_backbone}
\end{table}

\begin{figure}[t]
    \centering
    \vspace{-5mm}
\includegraphics[height=3cm]{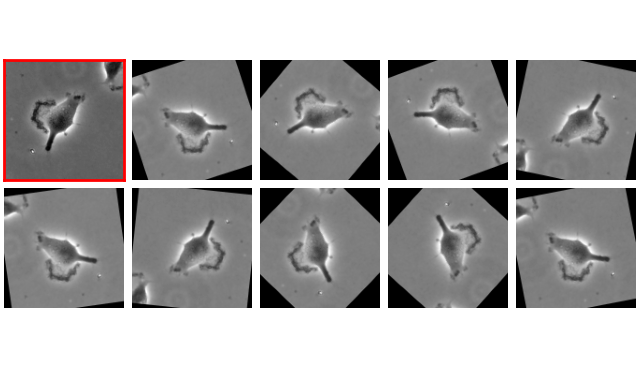}
    \vspace{-7mm}
    \caption{TTA for migration direction estimation ($n=10$).}
    \label{fig:ensemble}
\end{figure}

\vspace{1mm}
\noindent{\bf Test-time augmentation (TTA).}
Since SIECMD is a challenging task, we also investigate the potential of TTA for an ensemble effect \cite{Oza2024,Xu2022_MICCAI}. To achieve this, multiple rotated versions of each test cell image are generated. The migration direction estimated for each rotated image is then corrected by the applied rotation. For a total of $n$ cell images (one original and $n-1$ rotated versions), we obtain $n$ migration direction angles (see Figure \ref{fig:ensemble} for an example), which are fused to produce the final result, see \cite{Bruns2024} for details on the fusion method.

\vspace{1mm}
\noindent{\bf Estimation error evaluation and comparison with \cite{Nishimoto2019}.}
After training, we evaluate the migration direction estimation error on the test data (of size $N$) by calculating the mean deviation across all predictions $\alpha_i$ and their corresponding ground truth $\beta_i$:
\begin{eqnarray}
    \mathbf{E}_{deg} &=& \frac{1}{N} \sum_{i=1}^N \min(|\alpha_i - \beta_i|, \ 2\pi - |\alpha_i - \beta_i|)
    \label{eq_metric} 
\end{eqnarray}

In the related work \cite{Nishimoto2019}, the SIECMD task is framed as a 4-class classification problem (one class for each quadrant). For performance comparison, we estimate the mean deviation (in degrees). One approach is to compute the mean value between the minimal and maximal deviation:
\begin{eqnarray}
    \mathbf{E}_{deg} &=& \frac{1}{N} \sum_{i=1}^N \frac{\mbox{min}\_\mbox{deviation}_i + \mbox{max}\_\mbox{deviation}_i}{2} 
    \label{eq:mean_deviation_bin}
\end{eqnarray}
The maximal deviation can be minimized by treating the prediction as the central angle of the class (quadrant). For example, a ground truth class "1st quadrant" corresponds to $45^\circ$. The mean deviation for a correct prediction in the 1st quadrant is calculated as $(0 + \frac{\pi}{4})/2 = 22.5^\circ$. If the predicted quadrant is adjacent to the ground truth (2nd or 4th quadrant), the mean deviation is $(\frac{\pi}{4} + \frac{3\pi}{4})/2 = 90^\circ$. For the opposite (3rd) quadrant, the mean deviation is $(\frac{3\pi}{4} + \pi)/2 = 157.5^\circ$.

When comparing the mean deviation from \cite{Nishimoto2019} to our results, one issue is the lack of proportions for the three incorrectly predicted classes. To favor the external results, it is assumed that all incorrect predictions are off by only one quadrant. Given the classification accuracy $p$, the mean deviation (\ref{eq:mean_deviation_bin}) can then be estimated by:
\begin{eqnarray}
    \mathbf{E}_{deg} \ = \ p \times 22.5^\circ + (1-p) \times 90^\circ
    \label{eq:mean_deviation_bin_plos}
\end{eqnarray}
Note that this is an optimistic estimate, and the true $\mathbf{E}_{deg}$ is generally higher.

\section{Experimental validation}
\label{sec:results}

\noindent{\bf Data.}
Our performance evaluation is based on two datasets used in \cite{Nishimoto2019} and one additional dataset. These datasets contain time-lapse phase-contrast microscopic images of cell migration, captured in different ways.
%
\begin{itemize}
\item NIH3T3.
This cell line was isolated from a mouse embryo and frequently studied for its cellular migration patterns. To generate the dataset, the authors cultivated the cells in an on-stage incubation chamber under optimal growth conditions. Cellular movement was documented using phase-contrast microscopy images taken with a 20x objective at 5-minute intervals.
\item U373.
Phase-contrast microscopy images of the glioblastoma astrocytoma cell line U373 were previously used as a dataset for the ISBI cell tracking challenge in 2015. This cell line serves as a suitable model for studying the morphology and migration of cancer cells. In this context, it provides the opportunity to analyze changes in cellular behavior in response to their microenvironment or the influence of interfering drugs \cite{Ulman2017}.
\item MS3T3.
This dataset was generated at University XX using a similar NIH3T3 fibroblast cell line as in the NIH3T3 dataset published by Nishimoto \textit{et al.} \cite{Nishimoto2019}. Brightfield imaging was performed at 5-minute intervals over an incubation period of 100 minutes, resulting in 20 frames.
\end{itemize}
Example images can be found in Figure \ref{fig:examples}. The NIH3T3/U373/MS3T3 dataset contains 6969/5653/4284 images, respectively. Note that in \cite{Nishimoto2019}, results are also reported on a third dataset, hTERT-RPE1. Although this dataset is available to us, it was already prepared for the 4-class problem, making it impossible to obtain angle ground truth. Therefore, this dataset was not used in our study.

We multiplied the training data using augmentation. This includes a random combination of rotation $\in[0, 2\pi)$, $x$ and $y$ shift $\in[-0.2, 0.2]$ percent of the image size, scaling with factor $\in[-0.1, 0.1]$, and vertical and horizontal mirroring.

\vspace{1mm}
\noindent{\bf Ground truth generation.}
The ground truth for the presented datasets consists of angles $\alpha \in [0, 2\pi)$. Using TrackMate software \cite{trackmate}, a traveled path is obtained. The individual positions ${ ..., (x_i, y_i), (x_{i+1}, y_{i+1}), ... }$ of each cell are then used to calculate the direction of movement $\alpha$ and the distance traveled $\Delta$ between two consecutive images.


Cells change their direction at short time scales. Consequently, migration pattern are only directional over short periods of time, whereas cell migration over longer timescales resembles a random walk \cite{Begemann2009,Maiuri2015}. For that reason, selecting the proper time window is an important parameter that critically defines the feasibility of the problem. The net replacement thresholds in this work were chosen to be 10 $\mu$m (MS3T3) and 5 $\mu$m (NIH3T3, U373).

\begin{table}[t]
\centering
\begin{tabular}{l | c l l r r | c l l r r}
    \hline
    Dataset & Encoding & Activation & Loss & $\mathbf{E}_{deg}$ & $\pm$ & Encoding & Activation & Loss & $\mathbf{E}_{deg}$ & $\pm$ \\
    \hline\hline
    \multirow[t]{3}{*}{NIH3T3} & \multirow[t]{3}{*}{1N} & $\varphi_{cyclic}$ & $\delta_{cyclic}^2$ & 80.26 & 5.35 & \multirow[t]{3}{*}{2N} & $\varphi_{identity}$ & $\delta_{dist}^2$ & 34.29 & 2.20 \\
     &  & $\varphi_{cyclic}$ & $\delta_{cos}$ & 66.54 & 12.10 &  & $\varphi_{identity}$ & $\delta_{eucl}$ & 30.75 & 0.84 \\
     &  & $\varphi_{identity}$ & $\delta_{cos}$ & 59.91 & 12.82 &  & $\varphi_{sigmoid2d}$ & $\delta_{dist}^2$ & {\bf 30.51} & 0.86 \\ \hline
    \multirow[t]{3}{*}{U373} & \multirow[t]{3}{*}{1N} & $\varphi_{cyclic}$ & $\delta_{cyclic}^2$ & 31.44 & 2.62 & \multirow[t]{3}{*}{2N} & $\varphi_{identity}$ & $\delta_{dist}^2$ & 22.70 & 1.63 \\
     &  & $\varphi_{cyclic}$ & $\delta_{cos}$ & 33.79 & 3.76 &  & $\varphi_{identity}$ & $\delta_{eucl}$ & 39.20 & 29.41 \\
     &  & $\varphi_{identity}$ & $\delta_{cos}$ & 44.38 & 13.34 &  & $\varphi_{sigmoid2d}$ & $\delta_{dist}^2$ & {\bf 21.90} & 1.05 \\ \hline
    \multirow[t]{3}{*}{MS3T3} & \multirow[t]{3}{*}{1N} & $\varphi_{cyclic}$ & $\delta_{cyclic}^2$ & 69.46 & 2.55 & \multirow[t]{3}{*}{2N} & $\varphi_{identity}$ & $\delta_{dist}^2$ & 75.05 & 17.26  \\
     &  & $\varphi_{cyclic}$ & $\delta_{cos}$ & 65.41 & 2.80 &  & $\varphi_{sigmoid2d}$ & $\delta_{dist}^2$ & {\bf 56.68} & 1.42 \\
     &  & $\varphi_{identity}$ & $\delta_{cos}$ & 70.42 & 9.11 &  & $\varphi_{sigmoid2d}$ & $\delta_{eucl}$ & 57.17 & 2.40 \\ \hline
\end{tabular}
\caption{Migration direction estimation by the probing CNN. The column "Encoding" indicates one or two output neurons. The column "$\mathbf{E}_{deg}$" shows the mean angle deviation in degrees with the standard deviation in column "$\pm$".}
\label{tab:result_all_configurations}
\end{table}

\begin{table}[tb]
\vspace{-4mm}
\centering
\begin{tabular}{l r r r r}
        \hline
        Dataset & Backbone & Pre-training & $\mathbf{E}_{deg}$ & $\pm$ \\
        \hline\hline
        \multirow[t]{3}{*}{NIH3T3} & YOLOv8 & COCO & 23.77 & 0.64 \\
         & EfficientNetV2 & ImageNet & {\bf 17.27} & 2.26 \\
         & Resnet50  & ImageNet & 20.07 & 0.53 \\ 
        \hline
        \multirow[t]{3}{*}{U373} & YOLOv8  & COCO & 18.19 & 2.73 \\
         & EfficientNetV2  & ImageNet & {\bf 16.93} & 3.69 \\
         & Resnet50  & ImageNet & 23.77 & 4.59 \\
        \hline    
        \multirow[t]{3}{*}{MS3T3} & YOLOv8  & COCO & 31.87 & 5.58 \\
         & EfficientNetV2  & ImageNet & {\bf 30.54} & 5.21 \\
         & Resnet50  & ImageNet & 33.35 & 3.04 \\ \hline
    \end{tabular}
\caption{Migration direction estimation by large neural networks using the optimal configuration derived from Table \ref{tab:result_all_configurations}.}
\label{tab:result_optimal_configuration}
\end{table}

\begin{table}[!htb]
\centering
\begin{tabular}{l l r r | l l r r | l l r r}
    \hline
    Dataset & TTA & $\mathbf{E}_{deg}$ & $\pm$ & Dataset & TTA & $\mathbf{E}_{deg}$ & $\pm$ & Dataset & TTA & $\mathbf{E}_{deg}$ & $\pm$ \\
    \hline\hline
    \multirow[t]{5}{*}{NIH3T3} & No TTA & 17.27 & 2.26 & \multirow[t]{5}{*}{U373} & No TTA & 16.93 & 3.69 & \multirow[t]{5}{*}{MS3T3} & No TTA & 30.54 & 5.21 \\
     & + 1 rot. & 17.41 & 2.48 & & + 1 rot. & 17.59 & 2.95 & & + 1 rot. & 29.82 & 3.75 \\
     & + 5 rot. & 17.33 & 1.88 & & + 5 rot. & 17.77 & 2.53 & & + 5 rot. & 27.58 & 3.15\\
     & + 9 rot. & 17.24 & 2.15 & & + 9 rot. & 17.57 & 2.86 & & + 9 rot. & 27.57 & 3.61 \\
     & + 13 rot. & 17.18 & 2.17 & & + 13 rot. & 17.47 & 2.59 & & + 13 rot. & 26.77 & 3.33 \\ \hline
    \end{tabular}

\caption{Migration direction estimation by TTA (EfficientNet backbone, pre-training on ImageNet).}
\label{tab:result_ensemble}
\end{table}

\vspace{1mm}
\noindent{\bf Experimental results.}
We used a 4-fold validation, where each fold was randomly generated and split into 40\% training, 10\% validation, and 50\% test sets. The estimation error of the probing CNN is shown in Table \ref{tab:result_all_configurations}, which presents a selection of the 21 combinations (see \cite{Bruns2024} for the complete results). It turns out that for all three datasets, the optimal configuration is: circle encoding (2D), activation function $\varphi_{sigmoid2d}$, and loss function $\delta_{dist}^2$, as highlighted in bold.


We then used this optimal configuration to test the performance of three networks: YOLOv8, ResNet50, and EfficientNet. The results are given in Table \ref{tab:result_optimal_configuration}. The results show that the EfficientNetV2 architecture with a custom head trained on ImageNet achieved the best results across all datasets, as highlighted in bold. For NIH3T3 and U373, a mean angle deviation of approximately $17^{\circ}$ was achieved, while for MS3T3, the mean deviation was around $30^{\circ}$. This represents an improvement of up to $26^{\circ}$ compared to the probing results with the same parameter configuration.

Finally, using EfficientNetV2, we further studied the performance of TTA for $n = 2, 6, 10, 14$, as shown in Table \ref{tab:result_ensemble}. When comparing the results of $n = 1$ (no TTA applied) to $n = 14$, improvements were observed in both mean angle deviation and standard deviation for the NIH3T3 and MS3T3 datasets. For the U373 dataset, only an improvement in standard deviation was noted. When calculating the maximum estimation error (mean angle deviation + 3 $\times$ standard deviation), predictions for all three datasets benefited from TTA. Overall, it can be concluded that TTA has a positive effect on performance.

In MS3T3 cells, the direction estimation is less accurate compared to the other datasets. Considering reported cell-to-cell variability in persistence \cite{Maiuri2015}, we reason that variations in culturing and imaging conditions may reduce the duration of persistent migration, thus negatively impacting prediction accuracy.

\vspace{1mm}
\noindent{\bf Comparison with \cite{Nishimoto2019}.}
The major drawback of the only related work for solving SIECMD \cite{Nishimoto2019} is the coarse sampling of directions (on a unit circle) into just four discrete classes, which limits the directional resolution. A classification accuracy of $p = 87.89\%$ (NIH3T3) and $p = 81.76\%$ (U373) was reported in \cite{Nishimoto2019}, resulting in a mean deviation $\mathbf{E}_{deg}$ of 30.67$^\circ$ (NIH3T3) and 34.81$^\circ$ (U373) using the optimistic estimation (\ref{eq:mean_deviation_bin_plos}). These values are significantly higher than our results (17.27$^\circ$ for NIH3T3, 16.93$^\circ$ for U373). This comparison demonstrates considerable progress over the coarse classification approach in \cite{Nishimoto2019}.

\section{Conclusion}
\label{sec:conclusion}


In this paper, we addressed the problem of estimating cell migration direction from a single image. Our solution, using deep circular regression and cycle-sensitive methods, achieved an average error of about $17^\circ$ across two datasets, outperforming the previous approach. We also evaluated our method on a third, more challenging dataset. This work paves the way for previously impossible applications, with our future efforts focused on developing these applications.

%
%
%

\section*{Acknowledgments}
This work was supported by the Deutsche
Forschungsgemeinschaft (DFG): CRC 1450 – 431460824 (to X.J.), GA2268/3-1 (to M.G.), and GA2268/4-1 (to M.G.).

\bibliographystyle{splncs04}
\bibliography{bibliography}

\end{document}